\documentclass[letterpaper, 10 pt, conference]{ieeeconf}
\IEEEoverridecommandlockouts  
\let\labelindent\relax
\usepackage{enumitem}

\usepackage{mathtools,amsthm}

\usepackage{graphics} 
\usepackage{epsfig}
\usepackage{times} 
\usepackage{amsmath,amssymb,amsfonts}
\usepackage{algorithmic}
\usepackage{graphicx}
\usepackage{textcomp}
\usepackage{varioref}
\usepackage{import}
\usepackage{framed,xcolor}
\usepackage{color}
\usepackage{comment}
\usepackage{multirow}
\usepackage{epstopdf}   
\usepackage{psfrag}
\usepackage{breqn}
\usepackage{float}
\usepackage{mathtools}
\usepackage{booktabs}
\usepackage{soul}
\usepackage{amsbsy}
\usepackage[bottom]{footmisc}
\usepackage{lineno}
\usepackage{cleveref}
\usepackage{microtype}
\usepackage{comment}

\definecolor{arash}{rgb}{0.8,0.8,1}
\definecolor{seb}{rgb}{0.8,1,0.8}
\definecolor{seb2}{rgb}{0.5,.5,1}
\definecolor{arash2}{rgb}{0,.5,0}
\definecolor{wenqi}{rgb}{1,.75,0.79}
\definecolor{wenqi2}{rgb}{1,.75,0.79}

\graphicspath{{Figures/}}
\newcommand{\vect}[1]{\ensuremath{\boldsymbol{\mathrm{#1}}}}
\usepackage{graphicx}
\usepackage{tabularx,booktabs}
\usepackage{color}
\usepackage{multicol}
\usepackage{amsmath,amssymb}
\usepackage{amsfonts}
\usepackage{textcomp}
\usepackage{psfrag}
\usepackage{multimedia}
\usepackage{fancybox}
\usepackage{comment}
\usepackage{soul}

\definecolor{seb}{rgb}{0.8,1,0.8}
\definecolor{arash}{rgb}{0.8,0.8,1}

\newcommand {\matr}[2]{\left[\begin{array}{#1}#2\end{array}\right]}

\newcounter{lastnote}

\title{MPC-based Reinforcement Learning for Economic Problems with Application to Battery Storage} 
\author{Arash Bahari Kordabad, Wenqi Cai, Sebastien Gros
\thanks{The authors are with Department of Engineering Cybernetics, Norwegian University of Science and Technology (NTNU), Trondheim, Norway. E-mail:{\tt\small\{Arash.b.kordabad,wenqi.cai,sebastien.gros\newline
\}@ntnu.no}}
 }
\begin{document}
\bstctlcite{IEEEexample:BSTcontrol}
\maketitle
\thispagestyle{empty}
\pagestyle{empty}
\begin{abstract}
In this paper, we are interested in optimal control problems with purely economic costs, which often yield optimal policies having a (nearly) bang-bang structure. We focus on policy approximations based on Model Predictive Control (MPC) and the use of the deterministic policy gradient method to optimize the MPC closed-loop performance in the presence of unmodelled stochasticity or model error. When the policy has a (nearly) bang-bang structure, we observe that the policy gradient method can struggle to produce meaningful steps in the policy parameters. To tackle this issue, we propose a homotopy strategy based on the interior-point method, providing a relaxation of the policy during the learning. We investigate a specific well-known battery storage problem, and show that the proposed method delivers a homogeneous and faster learning than a classical policy gradient approach.
\end{abstract}
\section{Introduction}
\par Making decisions for the energy system in the presence of different forms of uncertainty is the object of recent publications~\cite{gross2020using,gross2020stochastic}. In smart grids, the uncertainty mainly arises from the imperfect forecasts for the prices, demand, and power generation. Finding a policy minimizing the economic cost of operating the grid in the presence of these uncertainties is highly valuable~\cite{powell2015tutorial}. Economic costs for smart grids are linear, based on the difference between the profit made by selling electricity to the power grid, and the losses incurred from buying it~\cite{harsha2014optimal}.
\par Reinforcement Learning (RL) offers tools for tackling Markov Decision Processes (MDP) without having an accurate knowledge of the probability distribution underlying the state transition~\cite{sutton2018reinforcement,bertsekas2019reinforcement}. RL seeks to optimize the parameters underlying a given policy in view of minimizing the expected discounted sum of a given baseline stage cost $L(\vect s,\vect a)\in \mathbb{R}$, where $\vect s,\vect a$ are the system states and inputs. 
 RL methods are usually either directly based on an approximation of the optimal policy or indirectly based on an approximation of the action-value function. Policy gradient methods directly seek to find the optimal policy parameters~\cite{SuttonPG,silver2014deterministic}. Different variants of Temporal Difference (TD) methods are at the core of many RL techniques for estimating the different value functions associated to the MDP. Least-Squares Temporal-Difference (LSTD) techniques are widely used because of their reliability and efficient use of data~\cite{lagoudakis2003least}.
\par Model Predictive Control (MPC) is a control strategy that employs a (possibly inaccurate) model of the real system dynamics to produce an input-state sequence over a given finite horizon such that the resulting predicted state trajectory minimizes a given cost function while explicitly enforcing the input-state constraints imposed on the system trajectories~\cite{rawlings2017model}. The problem is solved at each time instant, and only the first input of the input sequence is applied on the real system. By solving the entire problem at each time instant based on the current state of the system in a receding-horizon fashion, MPC delivers a policy for the real system. 
\par For computational reasons, simple models are usually preferred in the MPC scheme. Hence, the MPC model often does not have the structure required to correctly capture the real system dynamics and stochasticity. As a result, MPC usually delivers a reasonable but suboptimal approximation of the optimal policy. Choosing the MPC parameters that maximises the closed-loop performance for the selected MPC formulation is a difficult problem. Indeed, e.g. selecting the MPC model parameters that best fit the model to the real system is not guaranteed to yield the best closed-loop performance that the MPC scheme can achieve \cite{gros2019data}. In \cite{gros2020reinforcement,gros2019data}, it is shown that adjusting the MPC model, cost and constraints can be beneficial to achieve the best closed-loop performances, and RL is proposed as a possible approach to perform that adjustment in practice. Further recent research have focused on MPC-based policy approximation for RL~\cite{koller2018learning,bahari2021reinforcement,nejatbakhsh2021reinforcement,gros2020reinforcement}. \par MPC is a promising choice for the management of smart grids~\cite{gross2020stochastic}, because it provides a simple way to exploit forecasts on the grid prices, local power demand, and production, while respecting the physical limitations of the system. The stochasticity of the forecasts uncertainty is, however, not straightforward to treat at low computational costs. In this paper, we investigate a simple, well-known battery storage problem having a purely economic cost and stochastic dynamics. This example is has an optimal policy with a nearly bang-bang structure \cite{lifshitz2015optimal}, in the sense that the optimal policy selects inputs that are either in the bounds or zero for a large subset of the state space. We show that the deterministic policy gradient method is difficult to use for this type of problem because the state trajectories mostly lie in the set where the policy is trivially zero or in the bounds, which impedes the learning. 

\par In this paper we propose a homotopy strategy based on the interior-point method~\cite{biegler2010nonlinear}, which smoothens the MPC policy via the barrier parameter associated to the method, allowing for a more homogeneous and faster learning. The policy smoothing is gradually removed over the learning to recover the optimal policy. The paper is structured as follows. Section \ref{sec:model} presents the battery storage dynamics and provides its optimal policy of an economic cost. Section \ref{sec:RL} formulates the LSTD-based deterministic policy gradient method. Section \ref{sec:MPC} details the use of MPC-scheme as a function approximator in RL. The difficulties of applying the policy gradient method for (nearly) bang-bang policies is analyzed. And the main contribution of this paper is presented. Section \ref{sec:sim} provides the simulation results for the proposed approach and compares with the classical implementation of the policy gradient methods. Finally, section \ref{sec:conc} delivers the conclusions.
\section{A simple motivational example}\label{sec:model}
Photovoltaic (PV) battery systems allow households to participate in a more sustainable energy system~\cite{gross2020stochastic}. The local electric demand is covered by the PV battery system, or the connection to the public distribution grid. A simple model for the battery storage reads as \cite{gross2020using}:
\begin{align}
\label{eq:Dyna0}
\vect s_{k+1} &= \vect s _{k} + \alpha\left(\Delta_k +\vect a_k \right), 
\end{align}
where $\vect s_k\in[0,1]$ is the State-of-Charge (SOC) of the battery and the interval $[0,1]$ represents the SOC levels considered as non-damaging for the battery (typically 20\%-80\% range of the physical SOC). Constant $\alpha$ is a positive value that reflects the battery size. Variable $\Delta_k\sim\mathcal N\left(\bar \delta^X, \sigma^X\right)$ is the difference between the local power production and demand, which--for the sake of simplicity--is considered as a Normal centred random variable here, where $\bar\delta^X$ and $\sigma^X$ are the mean and variance of the Gaussian distribution. Input $\vect a_k\in [-\bar U,\bar U]$ is the power bought from (for $\vect a_k>0$) and sold to (for $\vect a_k<0$) the power grid. The economic stage cost can be written as follows:
\begin{align}
\label{eq:L}
L(\vect s_k,\vect a_k) = \left\{\begin{matrix}
\phi_b \vect a_k&\mathrm{if}&  \vect a_k \geq 0\\ 
\phi_s \vect a_k &\mathrm{if}&   \vect a_k < 0
\end{matrix}\right.,
\end{align}
where $\phi_b \geq 0$ is the buying price and $\phi_s\geq 0$ is the selling price, and we assume that $\phi_b \geq \phi_s$. For the sake of simplicity, 
we consider the prices $\phi_b$ and $\phi_s$ as constants. 
Appendix \ref{app:1} provides the model parameters we use in this paper. More complex models will be considered in the future.
\par In the deterministic policy gradient context, the optimal policy can be defined as follows:
\begin{align}
    \vect \pi^\star  = \mathrm{arg}\min_{\vect\pi}\mathbb E_{\vect\pi}\left[\sum_{k=0}^\infty \gamma^k \tilde L\left( \vect s_k,\vect a_k\right)\Bigg|\vect a_k=\vect \pi(\vect s_k)\right],
\end{align}
where $\gamma \in (0,1]$ is the discount factor, and for the battery storage dynamics \eqref{eq:Dyna0} with stage cost \eqref{eq:L}, the modified stage cost $\tilde L(\vect s_k,\vect a_k)$ is defined as follows~\cite{silver2014deterministic}:
\begin{align}
\label{eq:rl cost}
\tilde L(\vect s_k,\vect a_k) = L(\vect s_k,\vect a_k)&+p\max(\vect s_k-1,0)\nonumber\\&+p\max(-\vect s_k,0),
\end{align}
where $p$ is a large constant. The expected value $\mathbb E_{\vect\pi}$ is taken over the Markov Chain distribution resulting from the real system in closed-loop with policy $\vect\pi$. Since the state $\vect s_k$ ought to stay in the interval $\left[0,1\right]$, a large penalty is introduced in the RL stage cost for $\vect s_k\notin\left[0,1\right]$.
\par
\begin{figure}[ht!]
\centering
\includegraphics[width=0.3\textwidth]{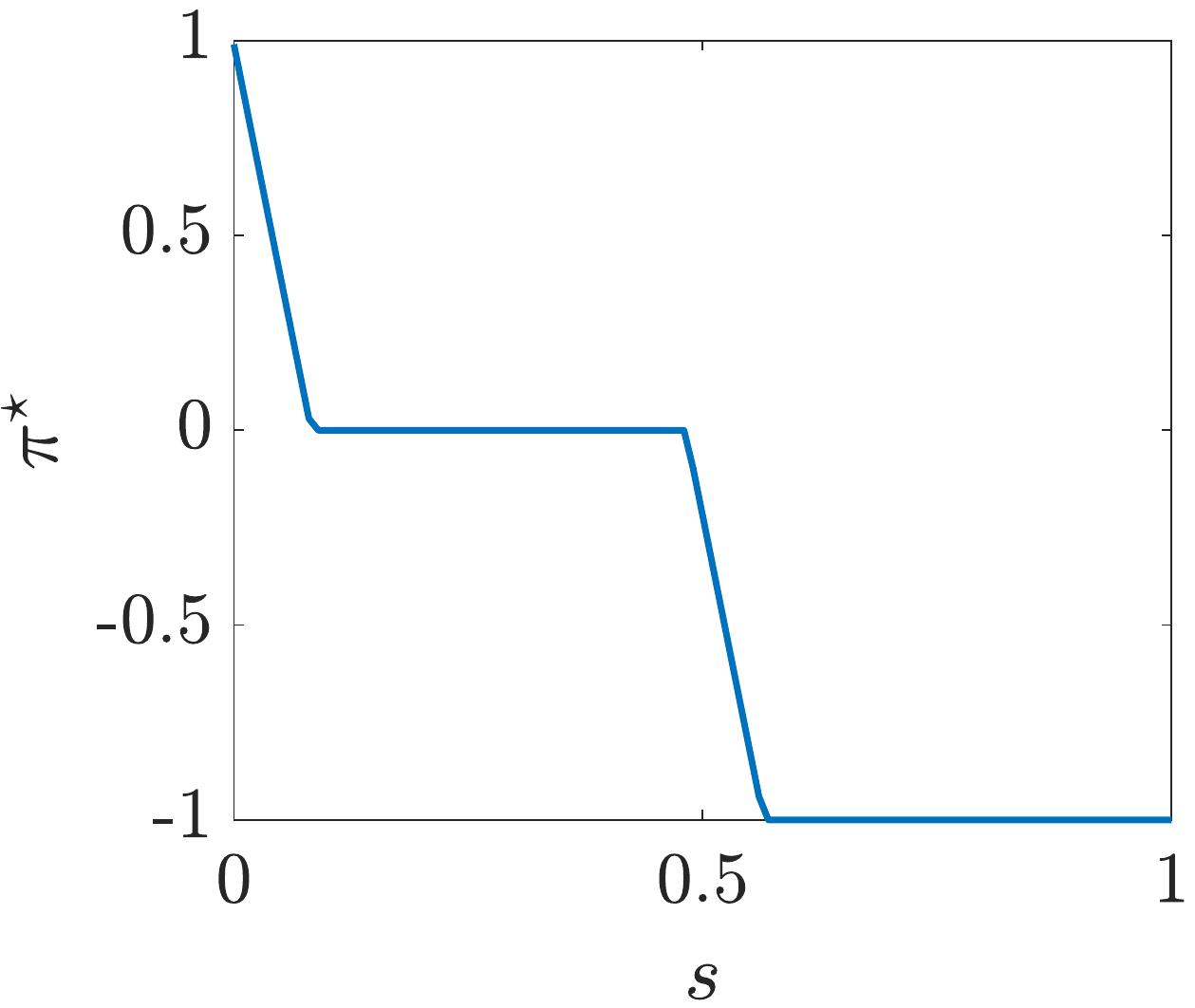}
\caption{Optimal policy $\vect\pi^\star$ resulted from DP.}
\label{fig:pi}
\end{figure}

 The example is selected such that its optimal policy $\vect \pi^\star$ can be solved via Dynamic Programming (DP), see fig. \ref{fig:pi}, and used as a baseline to assess the policies learned via RL. As can be seen in fig. \ref{fig:pi}, the optimal policy has a bang-bang-like structure. When the battery is at $\vect s\approx 0$, maximum buying is the optimal policy. Then for a fairly large subset of the states ($\vect s\approx [0.05,0.5]$), no exchange with the grid is the optimal policy. For a high SOC ( $\vect s\approx [0.55,1]$), maximum selling is optimum.
\par Note that the computational complexity makes DP unrealistic for systems more complex than this example. Instead, most investigations in RL (e.g., policy gradient methods) focus on achieving approximate solutions, which do not require a model of the dynamics. Next section details the RL algorithm that obtains an optimal policy based on the observed data from the (stochastic) real system.
\section{Deterministic policy gradient method}\label{sec:RL}
In the context of the deterministic policy gradient method \cite{silver2014deterministic}, the policy $\vect \pi_{\vect \theta}$ is parameterized by parameters $\vect \theta$, which are optimized directly according to the closed-loop performance using the gradient of the performance $J$ defined as:
\begin{align}
 J(\vect \pi_{\vect \theta}) = \mathbb E_{\vect \pi_{\vect \theta}}\left[\sum_{k=0}^\infty \gamma^k \tilde L\left( \vect s_k,\vect a_k\right)\Bigg|\vect a_k=\vect \pi_{\vect\theta}(\vect s_k)\right].
\end{align} 
The gradient of $J$ with respect to parameters $\vect\theta$ is obtained as follows:
\begin{align}\label{eq:dj}
    {\nabla _{\vect \theta} }J({\vect \pi _{\vect\theta} }) = {\mathbb E}_{\vect \pi_{\vect \theta}}\left[{\nabla _{\vect \theta} }{\vect \pi _{\vect \theta} }(\vect s){\nabla _{\vect a}}{A_{{\vect \pi _{\vect \theta} }}}(\vect s,\vect a)\big|_{\vect a=\vect \pi _{\vect \theta}}\right],
\end{align}
where $A_{{\vect{ \pi}}_{\vect{\theta}}}(\vect s,\vect a)=Q_{\vect{\pi}_{\vect{\theta}}}(\vect s,\vect a) - V_{\vect{\pi}_{\vect{\theta}}}(\vect s)$ is the advantage function associated to  ${\vect{ \pi}}_{\vect{\theta}}$, and where $Q_{\vect{\pi}_{\vect{\theta}}}$ and $V_{\vect{\pi}_{\vect{\theta}}}$ are the action-value and value functions for the policy ${\vect{ \pi}}_{\vect{\theta}}$, respectively. Under some conditions detailed in~\cite{silver2014deterministic}, the action-value function $Q_{{\vect \pi _{\vect \theta} }}$ in \eqref{eq:dj} can be replaced by an approximation ${Q_{\vect w}}$ without affecting the policy gradient. Such an approximation is labelled \textit{compatible} and can, e.g., take the form:
\begin{align}
\label{eq:Q_w}
    {Q_{\vect w}}\left(\vect s,\vect a\right ) = {\left(\vect a - {\vect \pi _{\vect \theta} }\left(\vect s\right )\right )^{\top}}{\nabla _{\vect \theta} }{\vect \pi _{\vect \theta} }{\left(\vect s\right )^{\top}}\vect w + {V_{\vect v}}\left(\vect s\right ),
\end{align}
where $\vect w$ is a parameters vector estimating the action-value function and $V_{\vect v}\approx V_{\vect\pi_{\vect \theta}}$ is a baseline function approximating the value function, which can, e.g., take a linear form:
\begin{align}
\label{eq:V_v}
    {V_{\vect v}}\left(\vect s\right ) =\Phi\left(\vect s \right)^\top {\vect v},
\end{align}
where $\Phi$ is a state feature vector and $\vect v$ is the corresponding parameters vector. The parameters $\vect w$ and $\vect v$ of the action-value function approximation \eqref{eq:Q_w} ought to be the solution of the Least Squares problem:
\begin{align}
\label{eq:error}
    \min_{ \vect w, \vect v} \mathbb{E} \left[\left( Q_{\vect \pi_{\vect\theta}}(\vect s,\vect a)-Q_{\vect w}(\vect s,\vect a)\right)^2\right].
\end{align}
In this paper, problem \eqref{eq:error} is tackled via Least Squares Temporal Difference (LSTD)~\cite{lagoudakis2003least}.
\par Next section details using an MPC scheme to approximate the optimal policy and proposes a smoothing approach based on the interior-point method for the (nearly) bang-bang policies. 
\section{MPC-based RL}\label{sec:MPC}
\par Using MPC as a way of supporting the approximations of value function, action-value function and policy $\vect \pi_{\vect \theta}$ has been proposed and justified in~\cite{gros2019data}. In this paper, we focus on the approximation of the optimal policy. Consider the following MPC scheme parameterized with $\vect \theta$:
\begin{subequations}
\label{eq:MPC1}
\begin{align}
 \min_{ \vect x, \vect u, \vect \sigma} \quad &T_{\vect \theta}\left( \vect x_N\right)+  \vect \omega_{\mathrm f}^\top \vect  \sigma_{N} \label{eq:mpc_cost}\\
&+\sum_{i=0}^{N-1}  \gamma^i(\ell_{\vect \theta}\left( \vect x_i,\vect u_i\right) +  \vect \omega^\top \vect \sigma_i)\nonumber \\
\label{eq:mpc_eq1}
\mathrm{s.t.}&\quad  \vect{x}_{i+1} = \vect f_{\vect\theta}(\vect{x}_{i},\vect u_i),\quad\vect x_0=\vect s\\
\label{eq:mpc_eq2}
&\quad \vect h_{\vect\theta}(\vect x_i,\vect u_i) \leq \vect \sigma_{i},\quad \vect h^{\mathrm f}_{\vect\theta}(\vect x_N) \leq \sigma_N \\
\label{eq:mpc_eq3}
&\quad \vect g(\vect u_i) \leq 0,\quad \vect \sigma_{i}\geq 0 , \quad \vect \sigma_{N}\geq 0,
\end{align}
\end{subequations}
where $T_{\vect\theta}$ and $\ell_{\vect\theta}$ are the MPC terminal and stage costs, respectively. Function $\vect f_{\vect\theta}$ is the model dynamics, $\vect g$ is the pure input constraint and $\vect{h}_{\vect\theta}$ and $\vect{h}_{\vect\theta}^{\mathrm f}$ are the stage and terminal inequality constraints, respectively.
Vectors ${\vect x}=\left\{\vect x_{0},\ldots ,\vect x_{N}\right\}$, ${\vect u}=\left\{\vect u_{0},\ldots,\vect u_{N-1}\right\}$ and ${\vect \sigma}=\left\{\vect \sigma_{0},\ldots ,\vect \sigma_{N}\right\}$ are the primal decision variables, $N$ is the prediction horizon and $\vect s$ is the current state of the system. Variables $\vect \sigma_{i}$ and $\vect \sigma_{N}$ are slacks for the relaxation of the state constraints, weighted by the positive vectors $\vect \omega$ and $\vect \omega_{\mathrm f}$. The relaxation prevents the infeasibility of the constraints of MPC in the presence of disturbances. The parameterized deterministic policy can be obtained as: 
 \begin{align}
    \vect \pi_{\vect \theta}(\vect s)=\vect u_{0}^{\star}(\vect s,\vect \theta),
\end{align}
where $\vect u_{0}^{\star}(\vect s,\vect \theta)$ is the first element of ${\vect u}^\star$, which is the solution of the MPC scheme \eqref{eq:MPC1}.

Theoretically, under some assumptions detailed in~\cite{gros2019data}, if the parametrization is rich enough, the MPC scheme is capable of capturing the optimal policy $\vect \pi^\star$ in the presence of disturbances and model error \cite{gros2019data}. 

\subsection{Primal-dual interior-point method}
In the following, we will use the primal-dual interior-point method to solve the MPC scheme \eqref{eq:MPC1}. Let us cast \eqref{eq:MPC1} as the generic Nonlinear Program (NLP):
\begin{subequations}
\label{eq:NLP:Gen}
\begin{align}
\min_{\vect z}&\quad \Psi_{\vect\theta}\left(\vect z\right) \\
\mathrm{s.t.}&\quad \vect G_{\vect\theta}\left(\vect z,\vect s\right) = 0 \\
&\quad \vect H_{\vect\theta}\left(\vect z\right) \leq 0,
\end{align}
\end{subequations}
where $\vect z = \left\{\vect x,\vect u,\vect\sigma\right\}$, function $\Psi_{\vect\theta}$ gathers the cost of \eqref{eq:MPC1}, and $\vect G_{\vect\theta}$, $\vect H_{\vect\theta}$ are its equality and inequality constraints, respectively. We denote $\mathcal L_{\vect\theta}(\vect y) = \Psi_{\vect\theta} + \vect\lambda^\top\vect G_{\vect\theta} + \vect \mu^\top \vect H_{\vect\theta}$ as the Lagrange function associated to \eqref{eq:NLP:Gen}, where $\vect y=\{\vect z, \vect \lambda, \vect \mu\}$ is the primal-dual variables vector, and where $\vect \lambda$ and $\vect \mu$ are the dual variables corresponding to the equality and inequality constraints, respectively. The primal-dual interior-point method is then based on the relaxed Karush–Kuhn–Tucker (KKT) conditions associated to \eqref{eq:NLP:Gen} as follows:
\begin{align}
\label{eq:IPKKT0}
    \vect r\left(\vect y,\vect s,\vect \theta\right) = \matr{cc}{\nabla_{\vect z}\mathcal L_{\vect \theta}\left(\vect y\right) \\
    \vect G_{\vect\theta}\left(\vect z,\vect s \right) \\
    \mathrm{diag}(\vect\mu_\tau) \vect H_{\vect\theta}\left(\vect z \right) + \tau\vect 1 },
\end{align}
and we denote its primal-dual solution by $\vect y_\tau = \left\{\vect z_\tau, \vect \lambda_\tau ,\vect \mu_\tau\right\}$ for each $(\vect s, \vect\theta)$ pair, i.e:
\begin{align}
\label{eq:IPKKT}
    \vect r\left(\vect y_\tau,\vect s,\vect \theta\right) =0,
\end{align}
where $\tau$ is the barrier parameter associated to the primal-dual interior-point method. Operator ``diag" gathers the vector elements on the diagonal elements of a square matrix and $\vect 1$ is a vector with unit elements and suitable size. If satisfying the Linear Independence Constraint Qualification (LICQ) and the Second Order Sufficient Condition (SOSC), $\vect y_\tau$ approximates a local solution of \eqref{eq:NLP:Gen} at the order of $\mathcal O(\tau)$~\cite{biegler2010nonlinear}.
\subsection{Policy sensitivity}
The policy gradient method requires one to compute ${\nabla _{\vect \theta} }{\vect \pi _{\vect \theta} }{\left( \vect s \right)}$ for every state $\vect s$ encountered by the policy (see Eq. \eqref{eq:dj}). It is therefore crucial to be able to compute ${\nabla _{\vect \theta} }{\vect \pi _{\vect \theta} }$ from data efficiently. We ought to recall here that $\vect \pi _{\vect \theta}$ is given by the first element of the input profile included in $\vect z$, delivered by NLP \eqref{eq:NLP:Gen}. In this paper, we will replace that solution by its interior-point approximation $\vect z_\tau$. The problem of computing $\vect \pi _{\vect \theta}$ then becomes the problem of differentiating the parametric solution $\vect z_\tau\left(\vect s,\vect\theta\right)$ of \eqref{eq:IPKKT} with respect to $\vect\theta$. If the original NLP \eqref{eq:NLP:Gen} satisfies LICQ and SOSC, then the sensitivity of $\vect z_\tau$ is readily given by the Implicit Function Theorem, i.e.:
\begin{equation}\label{eq:IFT}
    \left(\frac{\partial\vect r}{\partial \vect y }\frac{\partial \vect y}{\partial \vect \theta} + \frac{\partial \vect r}{\partial\vect\theta}\right)\bigg|_{\vect y=\vect y_\tau}=0
\end{equation}
holds. The policy sensitivity ${\nabla _{\vect \theta} }{\vect \pi _{\vect \theta} }$ can then be extracted from \eqref{eq:IFT} as follows~\cite{gros2019data}:
\begin{align}
\label{eq:sensetivity}
{\nabla _{\vect \theta} }{\vect \pi _{\vect \theta} }\left(\vect  s \right) =  - {\nabla _{\vect\theta} }{\vect r }\left( {\vect y_\tau},\vect s,\vect\theta\right){\nabla _{\vect y}}{\vect r }{\left( {\vect y_\tau},\vect s,\vect\theta \right)^{ - 1}}\frac{\partial {\vect y}}{\partial {\vect u_0}}
\end{align}
\subsection{Smoothing strategy for (nearly) bang-bang policies}
\label{PG on bang-bang policies}
The solution of NLP \eqref{eq:NLP:Gen} can be seen as a function of the NLP parameters $\vect s,\vect \theta$, and can be a non-differentiable or even discontinuous function of $\vect s,\vect \theta$ when changes of active set occur. In that context, parameter $\tau$ acts as a ``smoothing" factor in the NLP solution, in the sense that for $\tau >0$, the parametric solution $\vect z_\tau\left(\vect s,\vect \theta\right)$ obtained from solving \eqref{eq:IPKKT} becomes a smooth function of $\vect s,\vect \theta$. For $\tau \rightarrow 0$, $\vect z_\tau$ tends asymptotically to the non-smooth solution of NLP \eqref{eq:NLP:Gen}, and the derivatives of $\vect z_\tau$ can become unbounded for some $\vect s,\vect \theta$. In contrast, for $\tau$ larger, all derivatives of $\vect z_\tau$ remain bounded, and of lower magnitudes. 

When the optimal policy has a (nearly) bang-bang structure--such as in the storage example investigated here--it is beneficial to adopt a policy approximation $\vect\pi_{\vect\theta}$ that approximates that structure well while remaining smooth, such that the policy gradient \eqref{eq:dj} is guaranteed to be valid. If such a policy approximation can be made arbitrarily close to the bang-bang structure, then \eqref{eq:dj} remains asymptotically well defined, and the approximation can approach the optimal policy. 

For non-episodic problems, such as the battery storage example considered here, the expected value operator $\mathbb E_{\vect\pi_{\vect\theta}}[.]$ in the policy gradient \eqref{eq:dj} is meant to be taken over the steady-state distribution of the Markov Chain resulting from applying the policy $\vect\pi_{\vect\theta}$ on the real system. If the MPC policy $\vect\pi_{\vect\theta}$ has a purely bang-bang structure meant to approximate $\vect\pi^\star$, for $\tau\rightarrow 0$, the interior-point policy approximation 
is asymptotically bang-bang. Then, the gradient of the policy $\nabla_{\vect\theta}\vect\pi_{\vect\theta}$, while remaining well-defined everywhere, tends to be nearly zero on large parts of the state space, and take very large (asymptotically infinite) values when the policy switches between the different input levels. Hence, while the policy gradient \eqref{eq:dj} remains formally correct, evaluating it via sampling the distribution of the Markov Chain becomes very difficult, because the set of states where $\nabla_{\vect\theta}\vect\pi_{\vect\theta} \approx 0$ has a measure close to unity, while $\nabla_{\vect\theta}\vect\pi_{\vect\theta}$ is very large on a set of very small measure. As a result, sample-based estimations of \eqref{eq:dj} have a very large variance, which impedes the learning. 

\par
\begin{figure}
  \centering
  \begin{tabular}{ c }
\includegraphics[width=0.46\textwidth]{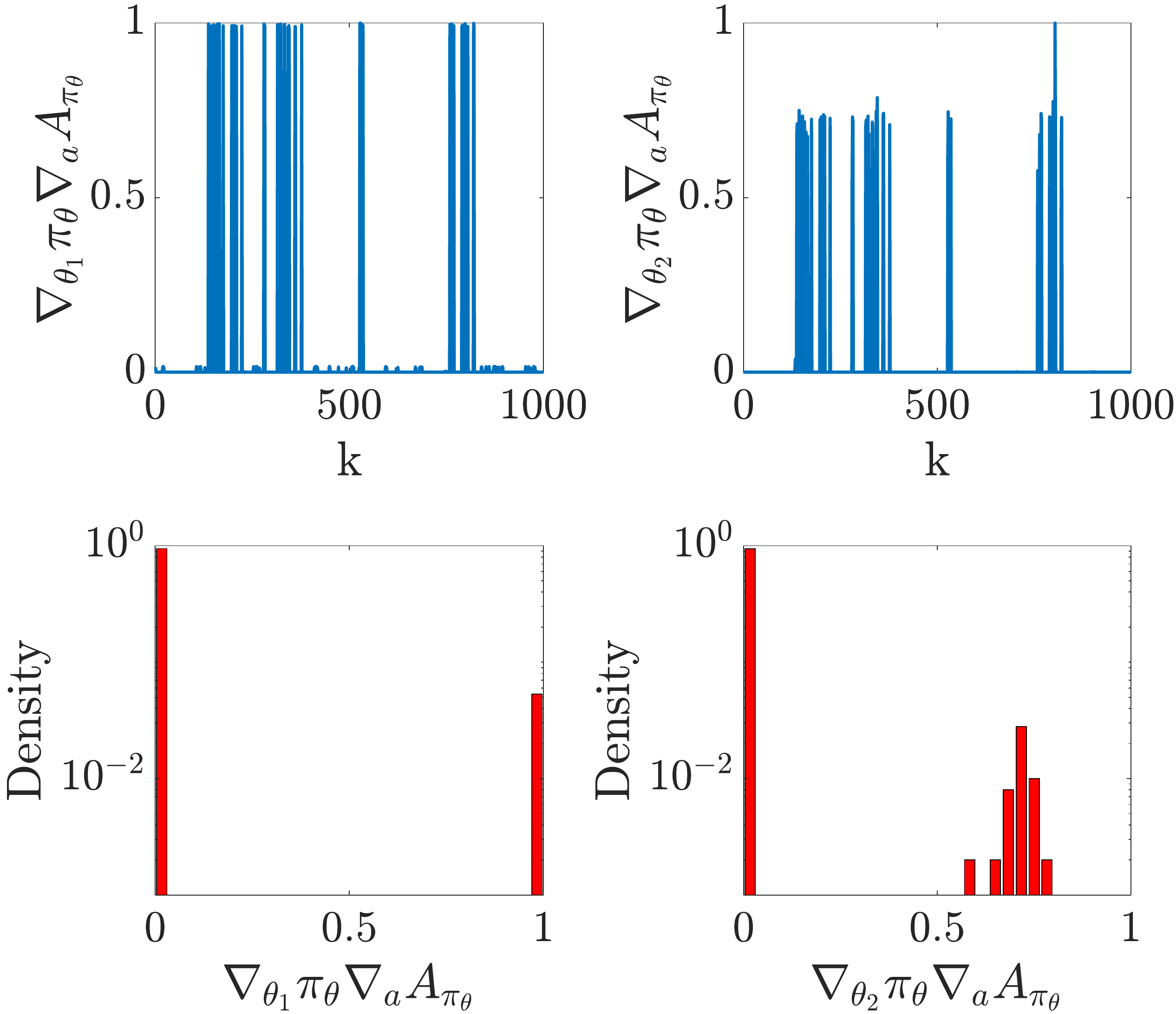}\\\small (a) $\tau=10^{-4}$\\
\includegraphics[width=0.46\textwidth]{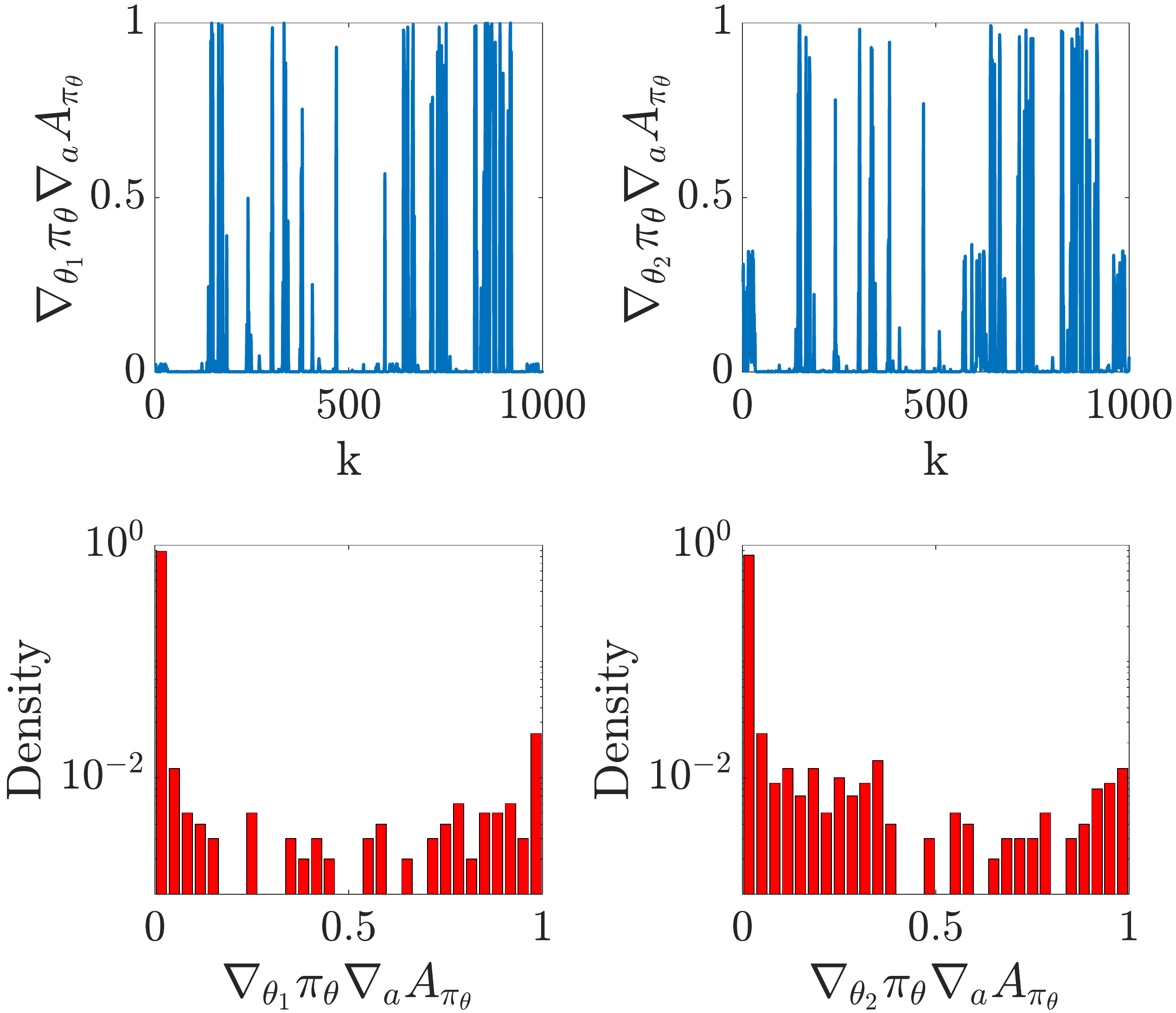} \\\small(b) $\tau=10^{-2}$
  \end{tabular}
  \medskip
\caption{Normalized $\nabla_{\vect\theta} \vect\pi_{\vect\theta} \nabla_{\vect a} A_{\vect\pi_{\vect\theta}}(\vect s,\vect a)|_{\vect a=\vect \pi_{\vect\theta}}$ with respect to $\theta_1$ and $\theta_2$ of a closed-loop trajectory for different values of $\tau$ and their densities. (The densities are in logarithmic scale.)}
\label{fig:tau}
\end{figure}

 For nearly bang-bang policy structures, the difficulties can be less severe than for purely bang-bang structure but they remain an issue. That issue can be observed for the battery storage problem considered in this paper. Figure \ref{fig:tau} shows the normalized $\nabla_{\vect\theta} \vect\pi_{\vect\theta} \nabla_{\vect a} A_{\vect\pi_{\vect\theta}}(\vect s,\vect a)|_{\vect a=\vect \pi_{\vect\theta}}$ for a given $\vect\pi_{\vect\theta}$ during a fairly long closed-loop trajectory for different values of $\tau$. Parameters $\vect\theta=[\theta_1,\theta_2]$ are the MPC parameters that will be introduced in detail in the simulation section. When using $\tau=10^{-4}$, it can be seen from Fig. \ref{fig:tau}(a) that the gradients are very close to zero for almost every time instance, while they are fairly large at some states $\vect s_k$ that are very close to the switching conditions in the bang-bang policy. This observation is clear in the density plot, where it can be seen that the value of the gradient is either zero or takes large values, without intermediate values. This indicates that during the learning, most of the time the policy gradient evaluation $\nabla_{\vect\theta}J$ is close to zero and takes large values when state trajectories yield large contributions $\nabla_{\vect\theta} \vect\pi_{\vect\theta} \nabla_{\vect a} A_{\vect\pi_{\vect\theta}}(\vect s,\vect a)|_{\vect a=\vect \pi_{\vect\theta}}$. In contrast, for results of $\tau=10^{-2}$ as displayed in Fig. \ref{fig:tau}(b), the distribution of the gradients is more uniform, avoiding the issues detailed above. 

\par
\begin{figure}
  \centering
  \begin{tabular}{ p{3.9cm} p{3.9cm} }
\includegraphics[width=0.22\textwidth]{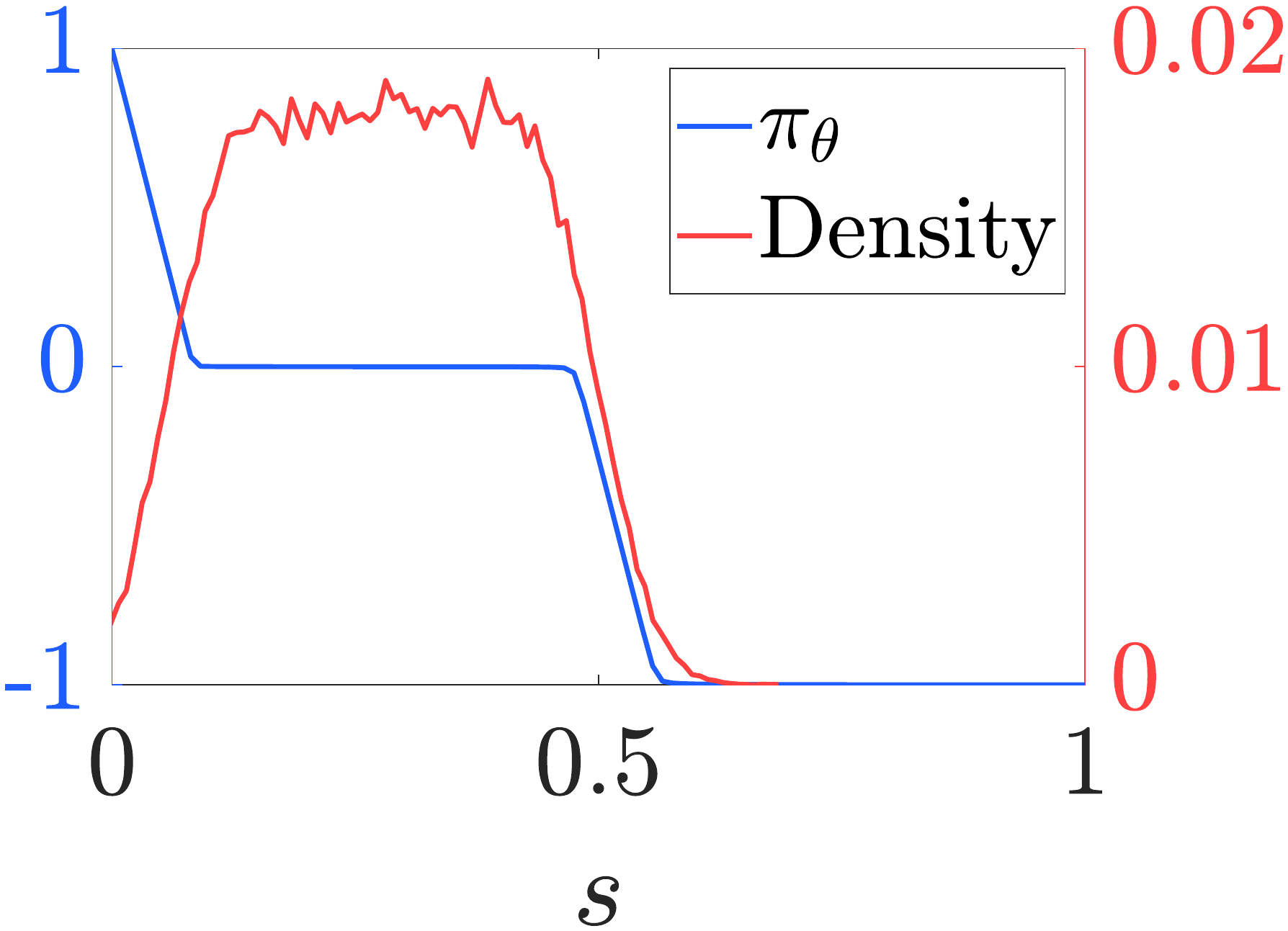} & \includegraphics[width=0.22\textwidth]{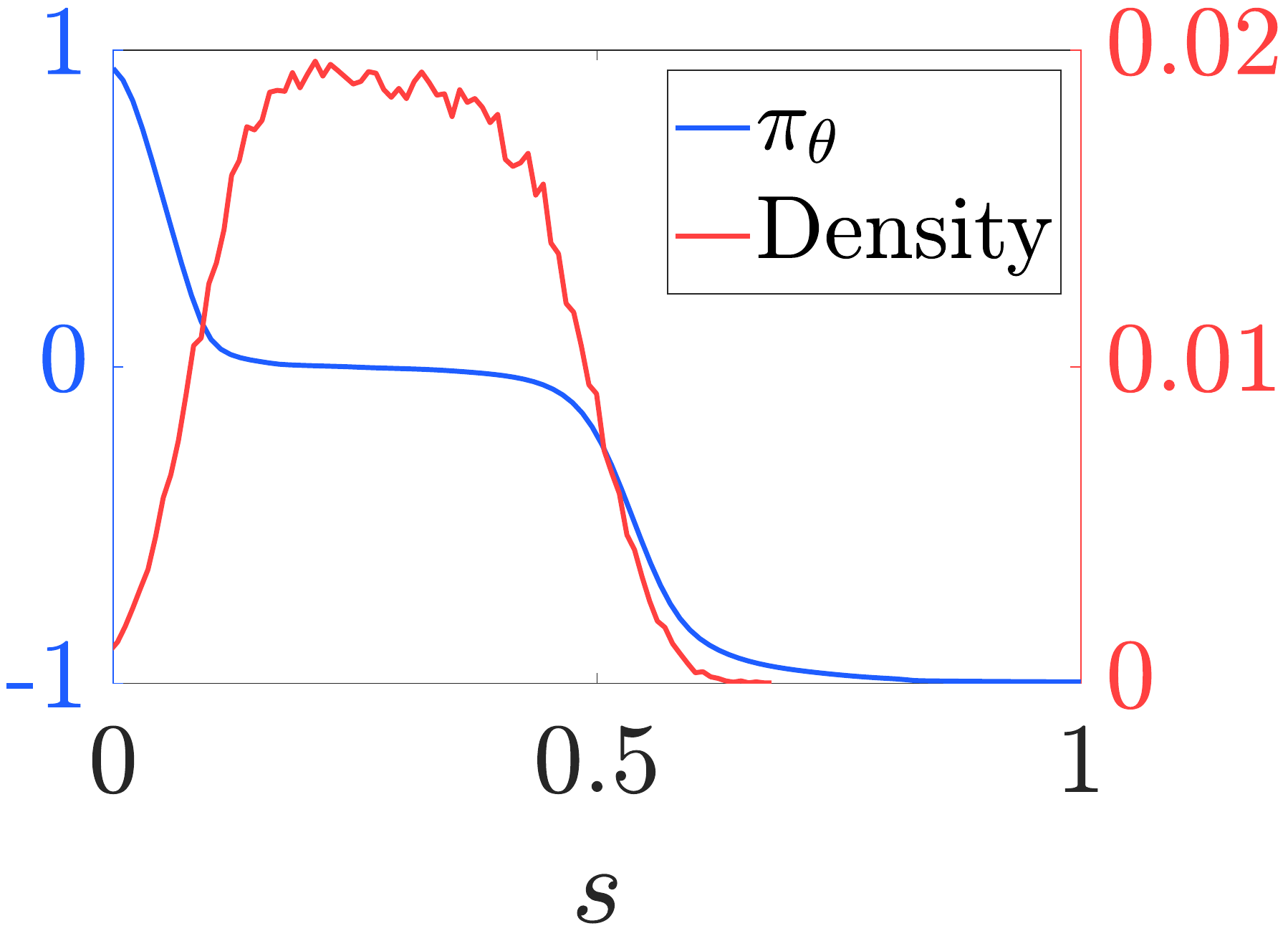}
\\\small(a) Non-smooth policy with state distribution for $\tau=10^{-4}$ & \small(b) Smooth policy with state distribution for $\tau=10^{-2}$
  \end{tabular}
  \medskip
\caption{MPC policy and state distribution of the closed-loop system}
\label{Fig.pimpc}
\end{figure}

 Figure \ref{Fig.pimpc} shows the MPC policy and the state distribution of the closed-loop system for two different values of $\tau$. It can be seen that for both $\tau$, the state density is mainly in the interval where the policy is trivially zero, hence the state trajectories rarely visit the set where $\nabla_{\vect\theta}\vect\pi_{\vect\theta}\neq 0$. Besides, for $\tau=10^{-4}$, the non-zero gradient occurs in a small subset of states. The policy $\vect\pi_{\vect\theta}$ tends to be the non-smooth solution of NLP \eqref{eq:NLP:Gen} and the values of $\nabla_{\vect\theta} \vect\pi_{\vect\theta} \nabla_{\vect a} A_{\vect\pi_{\vect\theta}}$ are relatively large for those data collected around the switching conditions. In contrast, for larger $\tau$ ($\tau=10^{-2}$), the policy $\vect\pi_{\vect\theta}$ is smoother. As a result, the values of sensitivity $\nabla_{\vect\theta} \vect\pi_{\vect\theta} \nabla_{\vect a} A_{\vect\pi_{\vect\theta}}$ remain bounded and with lower magnitude for large $\tau$ and they provide a meaningful gradient in a wider range of states compared with small $\tau$. 
\par In this paper, we exploit the smoothing effect of the barrier parameter $\tau$ to facilitate the use of the policy gradient method on MDPs that are difficult to treat because of the bang-bang-like structure of their optimal policy. More specifically, we propose to set the barrier parameter $\tau$ at large values the beginning of the learning to smoothen the policy and facilitate the learning, and decrease it--in a homotopy fashion-- to small values as the learning progresses towards the optimal policy. We adopt a linearly varying $\tau$ here that decreases from a large value to the targeted $\bar\tau$ of the interior-point method, i.e.
\begin{align}
\label{eq:taudecrease}
    \tau \leftarrow \max(\tau - \beta, \bar\tau)
\end{align}
where $\beta > 0$ is the progression step for $\tau$, and $\bar\tau$ the final barrier parameter targeted for the interior point method. The starting $\tau$ and target $\bar\tau$ are problem dependent. Alternative progression rules to \eqref{eq:taudecrease} can clearly be considered, including more advanced adaptive strategies.

\section{Simulation}\label{sec:sim}
In this section, we illustrate the difficulties encountered when using the LSTD-based policy gradient method to learn the nearly bang-bang optimal policy for the battery storage problem. We then demonstrate the proposed smoothing strategy as explained in section \ref{PG on bang-bang policies}. We ought to stress here that, this example has a policy that is not fully bang-bang, which allows the classical policy gradient method to work even without using the proposed technique. However, it requires significantly more RL steps and struggles with a high variance in the gradient estimation. A more extreme example with a pure bang-bang policy is likely to make the classic policy gradient method fail unless the proposed technique is used. Appendix \ref{app:1} gives the parameters of the model and RL used in the simulations. 

The explicit form of MPC scheme \eqref{eq:MPC1} used in the simulation is as follows:
\begin{subequations}
\label{eq:MPC_simulation}
\begin{align}
 \min_{ \vect x, \vect u, \vect \sigma} \quad &\theta_2^2(\vect x_{10}-0.5)^2 +  10\vect  \sigma_{10} \\
&+\sum_{i=0}^{9}  (0.99)^i(L\left(\vect x_i,\vect u_i\right) + 0.1\theta_1^2(\vect x_{i}-0.5)^2 +  10 \vect \sigma_i)\nonumber \\
\mathrm{s.t.}&\quad  \vect{x}_{i+1} = \vect{x}_{i} + 1/12 \vect u_i\\
&\quad \left[\begin{matrix}
\vect x_{i}-1\\-\vect x_{i} 
\end{matrix}\right] \leq \vect \sigma_{i},\quad \vect u_i \in [-1,1]\\
&\quad \vect x_0=\vect s,\quad \vect \sigma_{i}\geq 0 , \quad \vect \sigma_{10}\geq 0. 
\end{align}
\end{subequations}
We use quadratic stage and terminal costs with $0.5$ as their reference points. Parameters $\theta_1$ and $\theta_2$ tune the curvature of the costs, and are squared to ensure their positive definiteness, i.e. $\vect\theta:=[\theta_1,\theta_2]^\top$. Based on our simulation results, this parameterization is sufficient to capture the optimal policy.  

\begin{figure}[ht!]
\centering
\includegraphics[width=0.46\textwidth]{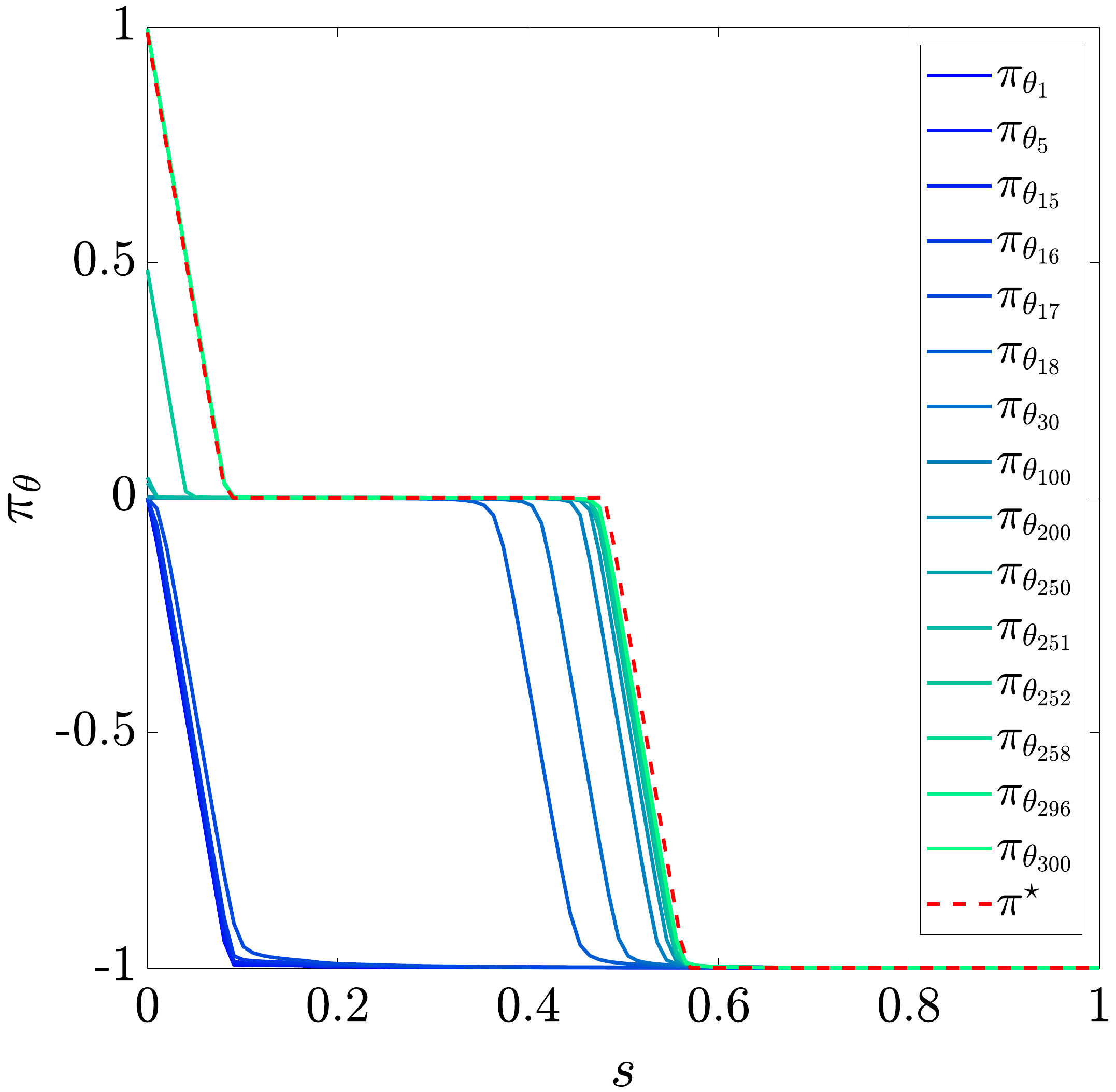}
\caption{The policy improvement process of the policy gradient method during $300$ steps with $\tau=10^{-4}$.}
\label{fig:PG300pi}
\end{figure}

\begin{figure}[ht!]
\centering
\includegraphics[width=0.48\textwidth]{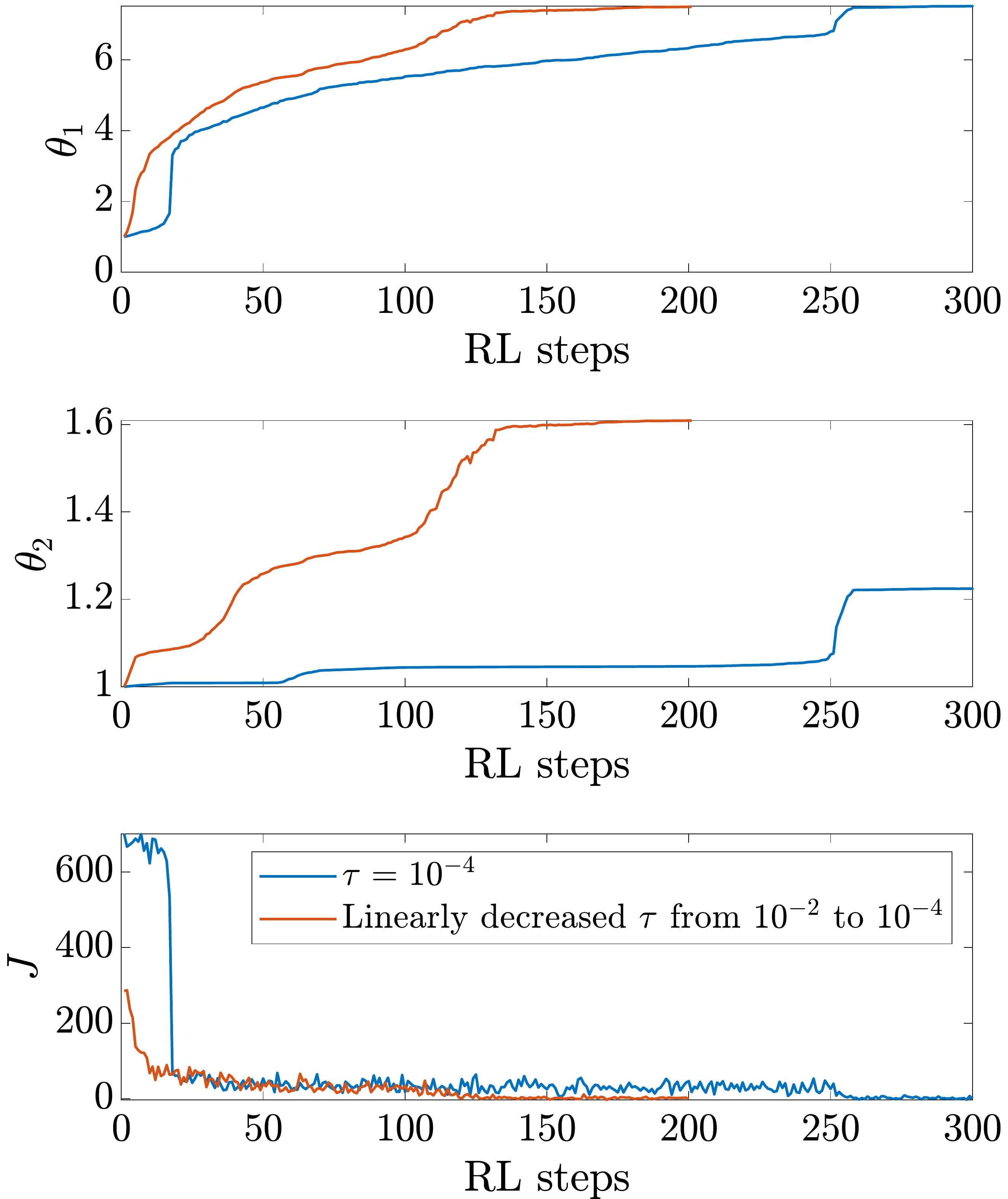}
\caption{The evolutions of the policy parameters $\theta_1$, $\theta_2$, and the closed-loop performance $J$ of the policy gradient method for the small $\tau=10^{-4}$ and linearly decreased $\tau$ from $10^{-2}$ to $10^{-4}$.}
\label{fig:thetaJforalltau}
\end{figure}

Figure \ref{fig:PG300pi} displays the policy improvement process for a fixed $\tau=10^{-4}$ using the LSTD-based policy gradient algorithm. Figure \ref{fig:thetaJforalltau} (blue curves) displays the policy parameters over the learning. One can observe that the learning progresses very slowly for long periods of time, when the state evolves in regions where $\nabla_{\vect\theta}\vect\pi_{\vect\theta_{k}}\approx 0$, and undergoes some infrequent, sudden changes otherwise. One can see in Fig. \ref{fig:PG300pi} that the policy gradient manages to approximate the optimal policy $\vect\pi^\star$ well, but the convergence is uneven.

\begin{figure}[ht!]
\centering
\includegraphics[width=0.46\textwidth]{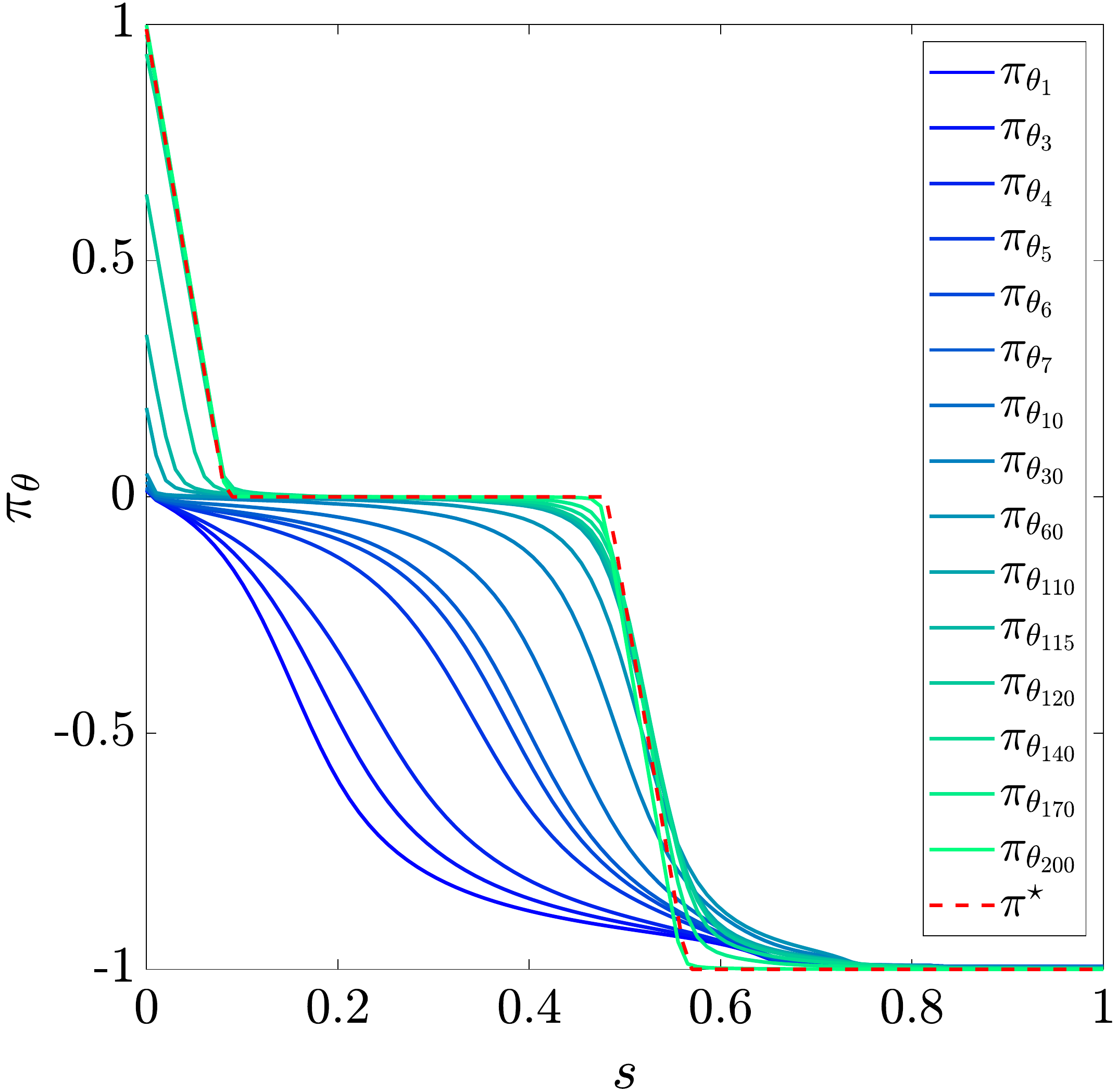}
\caption{The policy improvement process of the policy gradient method during $200$ steps with $\tau$ linearly decreases from $10^{-2}$ to $10^{-4}$.}
\label{fig:PGtaulinear}
\end{figure}

Figure \ref{fig:PGtaulinear} shows the policy improvement process resulting from $\tau$ starting at a relative large value and being progressively reduced to $\bar \tau$ using \eqref{eq:taudecrease}. The method starts with a large $\tau = 10^{-2}$, and the target $\bar\tau$ is $10^{-4}$. The step for decreasing $\tau$ is selected as $\beta = 5\cdot 10^{-5}$. With this choice of $\tau$, the policy is fairly smooth. The resulting learning can be seen in Fig. \ref{fig:thetaJforalltau} (light red curves). One can observe a significantly faster progression of the parameters, with a convergence in about 200 steps, as well as a significantly faster progression of the performance throughout the learning process, see Fig. \ref{fig:thetaJforalltau} lower graph. Starting with a larger $\tau$ when the optimal policy approximation is still inaccurate and reducing $\tau$ during the learning allows for a better learning progression and a better performance, while still delivering a policy having the correct structure because $\tau$ is reduced to a small value eventually. 

Hence, the proposed smoothing strategy not only accelerates the learning but also solves the dilemma between the smoothness of the policy improvement process and the accuracy of the obtained policy.
\section{CONCLUSION}\label{sec:conc}
In this paper, we discuss the use of the policy gradient method on policies having (nearly) bang-bang structures supported via MPC schemes. We detail why this kind of policy structure is difficult to treat in the deterministic policy gradient context, and propose a simple approach to alleviate the problem. A homotopy strategy is used to adapt the barrier parameter in the interior-point method that is used to solve the MPC scheme online. The proposed smoothing approach is illustrated on a classic battery storage problem with an economic stage cost. We show that a classical implementation of the policy gradient method results in a slow convergence, occurring through sudden progressions, while the proposed method offers a more homogeneous and faster convergence, resulting in a better closed-loop performance throughout the learning process. In future work, we will consider more sophisticated techniques to adapt the barrier parameter, analysize the convergence more formally, and tackle challenging economic problems with complex models.

\begin{appendices}

\section{Parameters of the dynamics and RL}\label{app:1}
\begin{table}[H]
\renewcommand\arraystretch{1.4}
\centering \scriptsize

\begin{tabular}{c|c|c}
\hline
\multirow{5}*{Dynamics} &$\phi_b$ & 5 \\ \cline{2-3}
 &$\phi_s$ & 2.5 \\ \cline{2-3} 
 &$\alpha$ & 1/12\\ \cline{2-3}
  &$\bar U$ & 1 \\ \cline{2-3}
 &$\Delta$ & $\mathcal N\left(0, 0.05\right)$\\\hline
 \multirow{2}*{RL} &$\Phi$ & $\left[\left( {s - 0.5} \right)^2,s,1\right]^\top$ \\\cline{2-3} 
 & $p$ & 1000\\\cline{2-3} 
 \hline
\end{tabular}
\end{table} 
\end{appendices}

\bibliographystyle{IEEEtran}
\bibliography{SGECC}
\end{document}